\documentclass[conference]{IEEEtran}

\usepackage[pdftex]{graphicx}
\usepackage{amsmath}
\usepackage{eqparbox}

\begin{document}

\title{\LARGE Semantic Segmentation on Multiple Visual Domains}
\author{\IEEEauthorblockN{Floris Naber}
\IEEEauthorblockA{June 2021\\
Department of Electrical Engineering\\
Eindhoven University of Technology\\
}}

\maketitle

\begin{abstract}
Semantic segmentation models only perform well on the domain they are trained on and datasets for training are scarce and often have a small label-spaces, because the pixel level annotations required are expensive to make. Thus training models on multiple existing domains is desired to increase the output label-space. Current research shows that there is potential to improve accuracy across datasets by using multi-domain training, but this has not yet been successfully extended to datasets of three different non-overlapping domains without manual labelling. In this paper a method for this is proposed for the datasets Cityscapes, SUIM and SUN RGB-D, by creating a label-space that spans all classes of the datasets. Duplicate classes are merged and discrepant granularity is solved by keeping classes separate. Results show that accuracy of the multi-domain model has higher accuracy than all baseline models together, if hardware performance is equalized, as resources are not limitless, showing that models benefit from additional data even from domains that have nothing in common.
\end{abstract}
\vspace{-0.3cm}
\section{Introduction} \label{sec:intro}
\begin{figure} [t]
    \centering
    \includegraphics[width=0.489\textwidth]{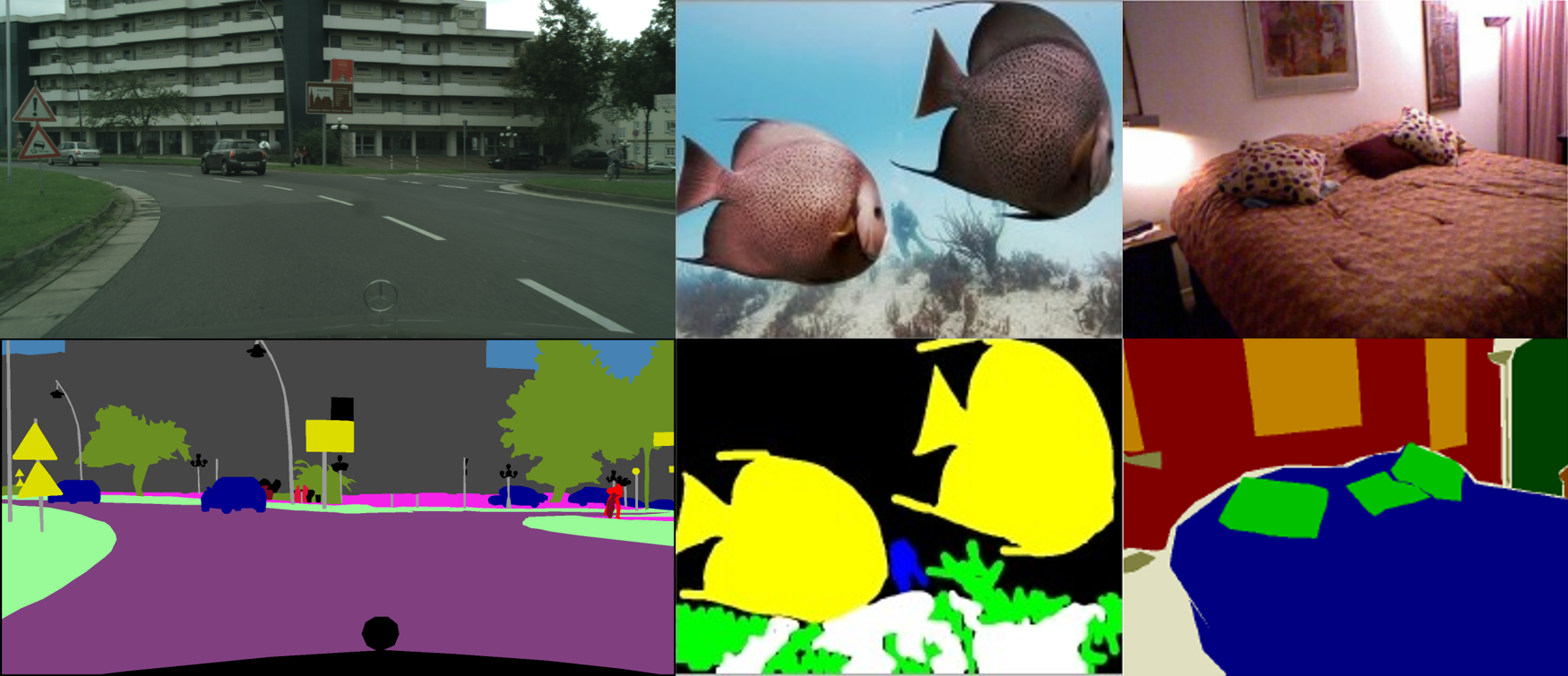}
    \vspace{-0.7cm}
    \caption{Example images (top) and colored annotations (bottom) of Cityscapes (left), SUIM (middle) and SUN RGB-D (right)}
    \vspace{-0.5cm}
    \label{fig:example}
\end{figure}

In the field of computer vision, one of the big challenges is semantic segmentation, which is the computer task of dividing an image into labelled regions that describe the contents of the image on a pixel level. These labels (or classes) could include anything, for example cars, people, chairs, trees and fish. Semantic segmentation has many diverse applications, like autonomous driving~\cite{Kaymak2018}, medical image analysis~\cite{Jiang2018}, infrastructure monitoring~\cite{Azimi2020}, robotics~\cite{Wolf2016}, photo editing~\cite{Gao2018} and much more. It is an especially important task in autonomous driving and robotics for example, because it is important for the models to understand the context in the environment in which they’re operating~\cite{Mwiti2019}.
Semantic segmentation can also be considered as a substantial preprocessing for others tasks, including object detection, instance segmentation and scene understanding~\cite{Saffar2018}.\\
Currently in the field, semantic segmentation is tackled with deep neural networks that are trained using supervised learning. This training is done using datasets consisting of images that are labelled with a set of predetermined labels, the so-called label-space. These datasets are labelled manually on a pixel level and are therefore very expensive to make, compared to for example datasets for object detection, which only require bounding box labelling.
As a result, only a limited amount of training data is available for semantic segmentation. Datasets are often small, have a limited domain and also a limited label-space, like urban driving scenes (Cityscapes~\cite{Cordts2016}), indoor scenes(SUN RGB-D~\cite{Song2015}) or underwater scenes (SUIM~\cite{Islam2020}) (see Fig. \ref{fig:example} for examples). This is undesired as neural networks thrive when they have access to a lot of diverse training data.\\ 
The accuracy of a model within a given domain can be very high, but as soon as a model leaves its domain, its performance can drop very quickly. For example a model trained on Cityscapes will perform well on urban roads, but will perform badly when facing a sidewalk with a restaurant and terrace, which is a situation better suited to a model trained on SUN RGB-D. While determining what image belongs to what dataset is very easy for datasets with a very clear domain, like the datasets mentioned above, this is not the case in the real world, as the border between different domains can be very blurry, as described by the example.\\
To improve accuracy on both domains, making a single model trained simultaneously on these different domains can improve the accuracy on both domains, compared to when both domains had different models trained for them specifically~\cite{Fourure2017}. Also in applications where a system needs to operate across multiple domains, it is desired that it only uses 1 model, as otherwise this burdens the developers with making multiple models for multiple domains. These models then have to run those simultaneously on multiple GPU's (expensive) or alternatively a controller needs to be developed to determine what model to use in every situation~\cite{Lambert2020,jain2020scaling}.\\
In this paper a method is proposed to combine three datasets from different domains (Cityscapes, SUN RGB-D and SUIM) without manual relabelling, to increase the size and diversity of the label-space a single model can train on. The goal is to research the hypothesis that model performance can be improved when trained on multiple domains, compared to models trained specifically for these domains combined. The system's hardware requirements during operation of the system should not be increased and the design should be simple enough to allow for inclusion of more datasets in the future without difficulty.\\
First the challenges of multi-domain training for the three used datasets are presented in Sec. \ref{sec:challenges} and in Sec. \ref{sec:related} the solutions to these challenges proposed by state-of-the-art research and their shortcomings are explained. In Sec. \ref{sec:method} a method that combines the three datasets is proposed that builds on the current research, after which the results are presented in Sec. \ref{sec:results}. The research is wrapped up with a discussion in Sec. \ref{sec:discussion} and a conclusion in Sec. \ref{sec:conclusion}.

\section{Challenges of multi-domain training} \label{sec:challenges}
Training a single neural network on multiple domains is not trivial, as there are challenges that need to be faced. The main challenge is the incompatibility between different datasets, as datasets often have incompatible label-spaces. Incompatibilities between the datasets used in this research consist of duplicate classes and discrepant granularity.\\
The first problem is when multiple datasets have a identical classes. For example, Cityscapes and SUN RGB-D both have a class \textit{person}, which are considered as separate classes if the datasets are merged without addressing this. This would increase the label-space size unnecessarily, which could impair prediction accuracy and could hinder model training, as it will try to learn differences between the 2 \textit{person} classes, while there are none in reality. For this reason a solution should be found such that the model sees these classes as identical.\\
The second problem of discrepant granularity is when something in an image corresponds to a specific class in one dataset, but multiple classes in another dataset. For example the class \textit{building} in Cityscapes, contains the classes \textit{door}, \textit{window} and \textit{wall} of SUN RGB-D. A simple solution to solve this issue would be to merge these more detailed classes of SUN RGB-D and map them to the Cityscapes \textit{building} class, but then detailed labelling that is available in SUN RGB-D would be lost, which is undesired. Also in an indoor domain it would be out of place to consider a \textit{building} class. A solution for this should be implemented that maintains a high granularity, while maintaining high accuracy.\\ 
\vspace{-0.3cm}
\section{Related work} \label{sec:related}
Currently, some research to address these incompatibility problems has already been done on multi-domain semantic segmentation. MSeg~\cite{Lambert2020} introduces a composite dataset made up of many popular datasets, including Cityscapes, Mapillary Vistas~\cite{Neuhold2017}, ADE20K~\cite{Zhou2017} and COCO~\cite{Lin2014}. The method for merging all the datasets is quite straight forward. A universal label-space was created by simply choosing one by hand based on all the included datasets. The total of 316 classes in all the datasets was reduced to 194 for MSeg, by merging most classes that were incompatible. To maintain a high granularity, some classes were manually relabelled to make those compatible with the rest of the dataset. The shortcomings of this idea are that even though a very detailed and diverse new dataset was created, still manual labelling is used to create a multi-domain dataset. Using this method, it is expensive to expand the label-space to new domains, like that of SUIM, which MSeg has a very bad performance on.\\
A semi-automated method of training a model on multiple incompatible datasets was proposed by Meletis and Dubbelman~\cite{Meletis2018, meletis2019data} using a hierarchical structure. Three datasets were used: Cityscapes, Mapillary Vistas and GTSDB~\cite{Ertler2020} (a bounding box traffic sign dataset). The solution proposed addresses the issue of discrepant granularity using a hierarchy, where the model first determines the regions in an image from the label-space of the highest hierarchy (i.e. lowest granularity). If on of these classes consists of multiple classes from a lower hierarchy, the model will relabel the pixels using the label-space of the lower hierarchy. A hierarchical structure makes for complex inference and is not computationally friendly, as the model has a head (part of the model that makes the predictions) for every hierarchical level. Also, this model did not expand the label-space to multiple domains, as every class in Cityscapes included in the final model corresponds to a class or a combination of classes from Mapillary Vistas, which means only the discrepant granularity problem within the driving domain is addressed. 
Bendavic et al.~\cite{Bevandic2020} proposed a method for creating a flat label-space for multiple domains, by creating a universal label-space that contains all the most detailed classes from all datasets included, after which all original classes are mapped to the classes of the universal label-space. Duplicate classes are simply mapped to the matching universal class and discrepant granularity is addressed by making the universal label-space match all the most detailed classes. So the Cityscapes \textit{road} is mapped to 8 more detailed universal road classes, which correspond to the road classes of Mapillary Vistas. This solution is very interesting, as the final model has a label-space with a higher granularity than all of the original datasets, but this method also makes the label-space very complicated, as many pixels end up having partial (or multiple) labels, which results in a dataset not having a defined ground truth for every pixels. As a result, the models trained with their universal label-space had varied results on different benchmarks, suggesting that a good solution for multi-domain semantic segmentation still is not achieved.\\
Multi-domain semantic segmentation is an issue that still has no good semi-automated solution. Research that currently has been done in the field still has shortcomings and did not manage to address all the problems. This research aims to build upon the works described above and explores a method of merging datasets from different domains effectively.

\section{Method} \label{sec:method}
In this section a method is proposed to merge 3 datasets; Cityscapes, SUN RGB-D and SUIM. First, the datasets are introduced in Sec. \ref{sec:choice}, secondly the dataset preparation is described in Sec. \ref{sec:prep}, then the merging method is described in Sec. \ref{sec:merge} and finally in Sec. \ref{sec:model} the model choice and optimization process is described. 
 \begin{figure*} [t]
    \vspace{-1cm}
    \centering
    \includegraphics[width=0.9\textwidth]{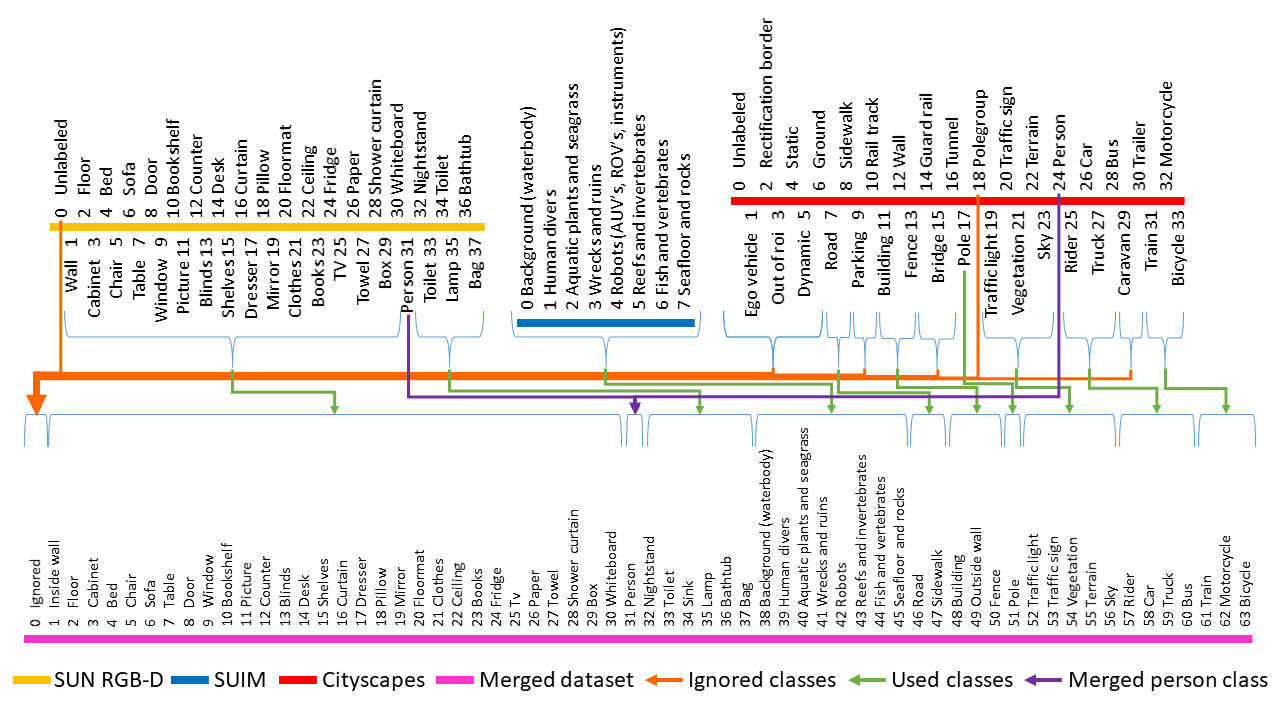}
    \vspace{-0.5cm}
    \caption{Class mapping of Cityscapes, SUIM and SUN RGB-D to the new universal label-space of the merged dataset.}
    \vspace{-0.5cm}
    \label{fig:idea}
\end{figure*}
 \subsection{Choosing datasets} \label{sec:choice}
 To achieve semantic segmentation on multiple domains effectively, datasets need to be chosen that match the criteria for this research.\\
 Firstly, the datasets should be from different domains and have useful classes, so a method of expanding the label-space and image-space can be researched. A useful class is one that a model can realistically encounter in a real life scenario. For example, brain tissue in Rontgen scans is not a useful class for a system that has regular camera images as input.\\
 Secondly, the datasets should have similar sizes, as solving problems regarding dataset size is not the focus of this research. For example, if a dataset has 10 times more images than another dataset, it would dominate the training data and therefor the model will be more suited to the domain of this dataset.\\
 Finally, the datasets should have a size and label-space large enough to achieve meaningful results and small enough to allow for more experimenting, as training on very large datasets with large label-spaces takes more time.\\
 The datasets that match all criteria and are chosen for this research are the Cityscapes dataset, containing 3475 training and validation images of urban driving scenes and 19 evaluation classes, SUN RGB-D, containing 5285 training and validation images of indoor scenes and 37 evaluation classes, and the SUIM dataset, containing 1525 train and validation images of underwater scenes and 8 classes. The complete label-spaces of the datasets can be seen in Fig. \ref{fig:idea}.
\subsection{Dataset preparation} \label{sec:prep}
To allow for easy training on all 3 datasets, dataset errors need to be removed and file types should be the same across datasets, so that model does have to deal with unnecessary differences during training.\\
Firstly, the annotation files of Cityscapes and SUIM are PNG-files and BMP-files respectively, while all the SUN RGB-D annotations are stored in a single matrix structure. To make the annotation file and type the same for all datasets, the SUN RGB-D annotation data was extracted and saved in PNG-files for every image. The binary color codes of the SUIM annotations where converted to integers and also stored in PNG-files.\\
Secondly, images and corresponding annotations that did not have matching sizes were found in SUIM and were removed from the dataset, reducing its size from 1525 to 1488 training and validation images.\\
Finally, SUIM and SUN RGB-D have no defined training and validation splits, while Cityscapes does. For SUN RGB-D and SUIM, 1000 and 292 images are chosen respectively for validation only, resulting in a roughly 1 to 5 split of training and validation data, which is similar to the Cityscapes split. The validation sets are chosen such that class representation per pixel was similar in the validation set and training set, to make sure the validation data is a fair representation of the training data.
\subsection{Merging the datasets} \label{sec:merge}
For this research it was chosen to make a merging method that is independent from the training and runs before training, so that no complex training or inference is necessary, to allow for further research with different models and implementation of more datasets and also to keep the system performance fast. It is also desired that the system maintains a high granularity, so the challenges stated in Sec. \ref{sec:challenges} should be addressed properly.
\paragraph{Duplicate classes}
This problem is addressed by mapping the duplicate classes to the same universal class. The \textit{person} class occurs both in Cityscapes and SUN RGB-D, as mentioned in Sec. \ref{sec:challenges}, so this class is mapped to the same class as SUN RGB-D \textit{person}. It should also be mentioned that both Cityscapes and SUN RGB-D have a class \textit{wall}, but these classes are unique. Cityscapes defines a wall as a standalone wall outside, not connected directly to a building, while SUN RGB-D defines walls as walls of a house. The classes have therefor been renamed accordingly, to textit{outside wall} and \textit{inside wall} to avoid confusion. 
Cityscapes classes which are typically ignored are mapped to the same class as SUN RGB-D \textit{unlabeled}, as these are all ignored during training and evaluation. SUIM does not have ignored classes. The mapping of every class to the universal label-space can be seen in Fig. \ref{fig:idea}.
\paragraph{Discrepant granularity}
A method that was researched to solve discrepant granularity is pseudo-labelling, which relabels the ground truth of a more general class like Cityscapes \textit{building} to more detailed classes like SUN RGB-D \textit{door}, \textit{window} and \textit{wall}, using a pretrained model. This would mean the final model trains on the predictions of another model, which is undesired as it reduces accuracy. hierarchical structures and partial labelling also were not chosen due to their complexity during inference.\\
The final solution that was chosen for this problem was to keep the labels mentioned above separate, as the classes occur in a very different context in the Cityscapes and SUN RGB-D domains and as it would result in the largest label-space. This way a wall, door or window seen from the outside is considered as part of the building, but on the inside it would be split up in the 3 different classes. Since all 4 classes have separate, non-overlapping definitions it should not cause any issues during training or testing.
\subsection{The model} \label{sec:model}
\paragraph{Choosing the model}
The focus of this research is not on developing a model so an existing model is chosen and slightly optimized for this research. Models used in this research are trained and optimized using the MMSegmentation toolbox~\cite{mmseg2020} of OpenMMLab. The model that was chosen for this research was a pyramid scheme parsing network, as this model was available with good hardware performance and thus allows for more experimenting. The specific model has lower accuracy than other more complex networks, but state-of-the-art accuracy is not the main goal of this research. 
\paragraph{Optimizing the model}
Bendavic et al.~\cite{Bevandic2020} state that multi-domain semantic segmentation benefits a lot from large batch sizes and crop sizes during training. For this reason it was chosen to downscale all the images to a size that would allow for large batch sizes, while not removing too much detail from the images, as this would decrease the accuracy of the model. An image resolution of 512x512 pixels was chosen, as this is was similar to the resolution of SUN RGB-D and SUIM images. These datasets have low resolutions and low quality annotations compared to Cityscapes, so a size that is similar to these datasets is the best compromise.\\
The crop size is chosen to be 256x256 pixels, as a smaller crop size was determined too small, because there would not be too little data in a single crop, which would result in the model not training effectively. Experimentation showed larger crop sizes resulted in over-fitting on the training set. During training crops are randomly taken from the images, so if the crop size is large, the randomness would decrease and risks of over-fitting increase.\\
The final image and crop sizes would allow for a batch size of 6 images per iteration on a desktop with a GTX 1070 with 8GB of Memory, which is the system used for experimenting.\\
The learning schedule that was chosen was a step function, which is independent from other training parameters like iteration total, as this allows for easy experimenting compared to other schedules. The final parameters that were set on was 80k iterations, meaning enough epochs would pass for the results to be useful. 80k iterations is quite a lot, but risk of over-fitting was very low due to the random crops. The learning rate would start at 0.01 and half every 20k iterations, which resulted in sufficient accuracy during validation.

\section{Results} \label{sec:results}
\subsection{Metrics}
To analyze the performance of the final models, the evaluation metric \textit{mean Intersection over Union} (mIoU) is used, which is currently the most used evaluation method in the field. Popular benchmarks use this metric to determine the order of the leaderboards. The main reason to choose this method over a method such as pixel accuracy (true positives per total amount of pixels) is because that method does not compensate for uncommon classes. mIoU on the other hand, takes the average of the IoU (true positives per all positives and false negatives) of every class~\cite{Tiu2019}, so accuracy on every class is equally important.
\subsection{Analysis} 
The final model will be compared to identical models trained on the same hardware setup. 3 models are trained on the datasets individually, to see how the accuracy of the model trained on the merged dataset compares to the accuracy of separate models. Such a system has several disadvantages as mentioned in Sec. \ref{sec:intro}, but it is a good baseline to compare with. Also a model is trained on Cityscapes and SUIM, as there is no discrepant granularity between these datasets, as it can then be observed if models trained on easily merged datasets improves accuracy compared to separate models.\\
Hardware performance of the baseline models combined is worse than the models trained on the merged datasets, as can be seen in table \ref{tab:specs}. The model trained on Cityscapes and SUIM performs best as expected, as the label-space is smaller than that of all 3 datasets combined, but this model does not work on the SUN RGB-D dataset, which can also be seen in Fig. \ref{fig:segm}.
\begin{table}[ht]
\vspace{-0.3cm}
\caption{Hardware performance of the baseline models added together, the model trained on Cityscapes and SUIM and the model trained on all 3 datasets merged.}
\centering 
\begin{tabular}{ c | c  c  c }
Train   & \shortstack{Training time\\1 iteration [ms]} & Memory [GB] & \shortstack{Output size\\ ~[\# labels]} \\ 
\hline   
Individual datasets & 450           & 14.7      & 19/8/37     \\  
Cityscapes + SUIM   & 149           & 4.9       & 27         \\ 
Citys + SUIM + SUN  & 180           & 4.9       & 63        \\  
\end{tabular} 
\label{tab:specs} 
\vspace{-0.3cm}
\end{table} 
When looking at the results from table \ref{tab:result}, it can be observed that the models trained on the individual datasets have the highest accuracy and the overall accuracy decreases when more datasets are merged. This can be explained by the fact that the individual models together have trained on 3 times more training data, as all models are trained with the same batch size and iteration total. If the batch size of the baseline models is changed such that the total amount of training data is the same for the baseline models together, then a fair comparison can be made. This is done in Sec. \ref{sec:batch}. The drop in accuracy from the baseline models to the model trained on Cityscapes and SUIM is smaller than the drop when all datasets are merged. This can be explained by the fact that SUN RGB-D has the largest label-space and image total, so it stands to reason that this dataset has the largest influence on the results.
\begin{table}[ht] 
\vspace{-0.3cm}
\caption{Semantic segmentation accuracy (mIoU [\%]) by models trained on the datasets Cityscapes, SUIM and SUN RGB-D individually, Cityscapes and SUIM merged, and all 3 datasets merged.}
\centering
\vspace{-0.2cm}
\begin{tabular}{ c | c  c   c }
                    & \multicolumn{3}{c}{Test}\\  
Train               & Cityscapes    & SUIM      & SUN RGB-D \\ 
\hline   
Individual datasets & 60.92         & 60.79     & 32.16     \\  
Cityscapes + SUIM   & 56.39         & 57.39     & 0.00      \\ 
Citys + SUIM + SUN  & 42.68         & 51.81     & 26.96     \\  
\end{tabular} 
\label{tab:result} 
\vspace{-0.4cm}
\end{table} 

\subsection{Accounting for batch size} \label{sec:batch}
\begin{figure} [t]
    \centering
    \includegraphics[width=0.489\textwidth]{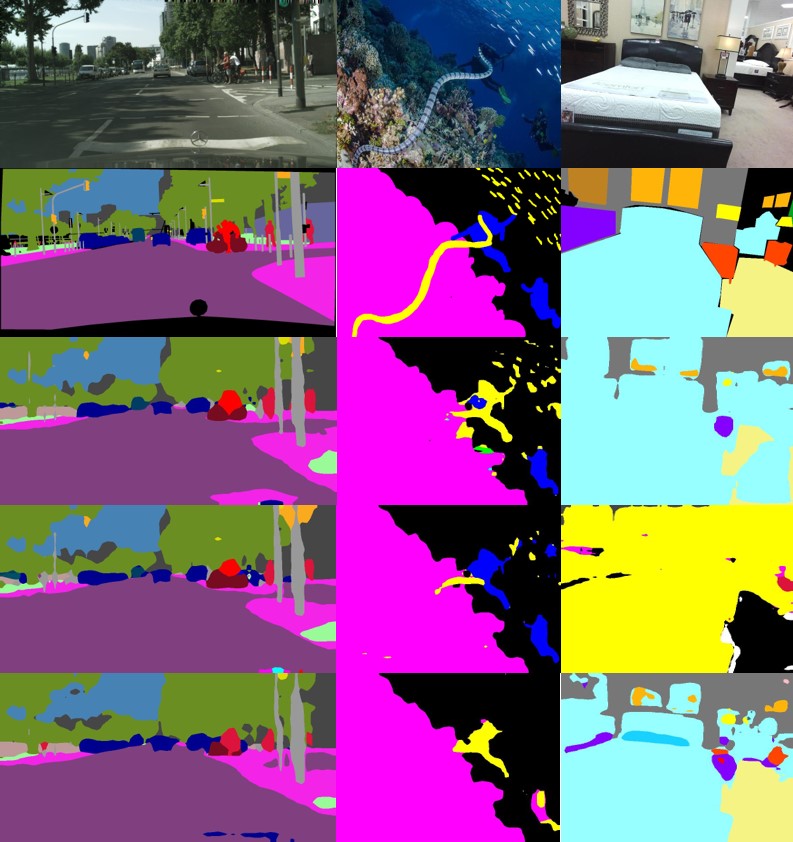}
    \vspace{-0.7cm}
    \caption{Semantic segmentation results on images (row 1) of Cityscapes (left), SUIM (middle) and SUN RGB-D (right) not seen by the models during training. Row 2 shows the ground truth, row 3 to row 5 show predictions by models trained on the individual datasets (batch size 3)(row 3), Cityscapes and SUIM (batch size 6)(row 4) and all 3 datasets (batch size 6)(row 5).}
    \vspace{-0.5cm}
    \label{fig:segm}
\end{figure}

If batch size is reduced from 6 to 3 for the baseline models of Cityscapes and SUIM, the image total and training time is the same as for the model trained on both of them. The results in this case (table \ref{tab:batch}) show that the model trained on both datasets has a higher accuracy on SUIM than the baseline model of SUIM, while the models have both trained on approximately the same amount of training data from this dataset. This improvement can also clearly be seen in the segmentation results in Fig. \ref{fig:segm}. The results of the model trained on both datasets still show worse results compared to the baseline model of Cityscapes. In Fig. \ref{fig:segm} and in table \ref{tab:batch} it can be seen that the difference is very minor. One of the causes could be that the SUIM dataset is too small to benefit a model trained on Cityscapes.\\
Also the batch size is reduced from 6 to 3 for the baseline model of SUN RGB-D, as this dataset comprises around half of the total amount of training data. Now the model trained on all 3 datasets scores an mIoU of 5\% more than the baseline model of SUN RGB-D, even though they have seen the same amount of training data from this domain. This improvement can very clearly be seen in the segmentation results in Fig. \ref{fig:segm}, showing that training on multiple domains improves the accuracy on this domain a lot.\\
The overall performance increase of the model trained on all 3 domains, compared to these baseline models, can be explained by the fact that the model is exposed to a larger variety of data and labels at the same time and therefor has a more detailed understanding of what features define each class and is therefore able to separate classes better, even within a single domain, than models trained on every domain separately.
\vspace{-0.3cm}
\begin{table}[ht] 
\caption{Semantic segmentation accuracy (mIoU [\%]) by models trained on the datasets individually with reduced batch size and models trained on merged datasets with original batch sizes.}
\vspace{-0.1cm}
\begin{tabular}{ cc | ccc } 
                    & Batch            & \multicolumn{3}{c}{Test}\\  
Train               & size    & Cityscapes    & SUIM  & SUN RGB-D \\ 
\hline   
Individual datasets & 3             & 57.45         & 55.02 & 22.12     \\  
Cityscapes + SUIM   & 6             & 56.39         & 57.39 & 0.00      \\
Citys + SUIM + SUN  & 6             & 42.68         & 51.81 & 26.96     \\
\end{tabular} 
\label{tab:batch}
\vspace{-0.7cm}
\end{table} 

\subsection{Discrepant granularity and duplicate classes} \label{sec:discr}
The merging of the \textit{person} classes showed a huge accuracy increase in the SUN RGB-D validation set, as can be seen in table \ref{tab:class}. The IoU is improved from 0.09\% to 2.65\%. The accuracy increase can be explained by the fact that the \textit{person} class is extremely rare in the SUN RGB-D dataset and therefor models trained on this dataset benefit a lot from additional training data containing this class more often. On the other hand it could also just be the case that this improvement is caused by adding more datasets to the training data, as the increase is in line with the average accuracy increase. There is an accuracy decrease of \textit{person} class in Cityscapes, which is less then the overall mIoU decrease of Cityscapes (see table \ref{tab:result}), suggesting that the merging does not contribute towards the mIoU drop.\\
The solution to discrepant granularity between Cityscapes and SUN RGB-D resulted in a decrease of around 5\% for all 4 involved classes when the datasets are merged, as can be seen in table \ref{tab:class}. This drop in accuracy is significantly less than the overall drop in mIoU on the domains of both Cityscapes and SUN RGB-D when the datasets are merged.\\
When looking at the baseline model with a reduced batch size, it can be seen that there is no performance drop in any of the SUN RGB-D classes when the datasets are merged. The accuracy on the class \textit{door} even increases by 2\%. An explanation for no accuracy drop could be that the context in which the conflicting classes occur is very different, namely in the Cityscapes and SUN RGB-D domain, and that the class definitions do not necessarily overlap. As a result the model can still distinguish these classes.
\begin{table}[ht] 
\vspace{-0.3cm}
\caption{Semantic segmentation accuracy (IoU [\%]) of 5 different classes by models trained on the datasets Cityscapes and SUN RGB-D individually and all 3 datasets merged together.}
\centering
\vspace{-0.2cm}
\begin{tabular}{ c | c  c  c | c c }
                    & \multicolumn{3}{c|}{Individual datasets}& \multicolumn{2}{c}{All 3 datasets}\\ 
Batch size          & 6             & 6         & 3     & \multicolumn{2}{c}{6}\\
\hline
\shortstack{\\Class}& \shortstack{City-\\scapes}    & \shortstack{SUN\\RGB-D} & \shortstack{SUN\\RGB-D} & \shortstack{City-\\scapes}& \shortstack{SUN\\RGB-D} \\
\hline   
Person              & 62.79         & 0.09      & 0.00  & 47.92     & 2.65  \\  
Building            & 86.61         & -         & -     & 82.08     & -     \\ 
Inside wall         & -             & 70.92     & 66.29 & -         & 65.94 \\
Door                & -             & 29.77     & 22.11 & -         & 24.48  \\
Window              & -             & 43.55     & 37.83 & -         & 38.08  \\
\end{tabular} 
\label{tab:class} 
\vspace{-0.4cm}
\end{table} 

\section{Discussion} \label{sec:discussion}
The discrepant granularity solution shows accuracy can be improved on classes when their definitions do not overlap and the context in which they occur is different. This solution will not work between datasets where the context is similar or where definitions clearly overlap. Further research can explore how to address this while keeping a high granularity and flat label-space to avoid training and inference complexities.\\
The results of this research could be further improved if hardware used during experimentation was more powerful, as batch sizes clearly has a large influence on results. Further research can explore how batch size and batch choice influences the performance of a model trained on multiple domains.\\ 
During this research it has also been found that a model trained on multiple non-overlapping domains is extremely good in learning the differences between individual domains. During evaluation, the final model will almost exclusively predict classes from the domain every image is from. Further research could dive into why this happens and if there is a way to use this phenomenon to improve model performance.\\
A dataset that was first considered, but ultimately not used in this research, is COCO-stuff, because it has too many labels and images, 182 labels and over 100,000 images. The diverse label-space resulted in very long training times and the system was very slow at handling the large amount of files contained in the dataset, due to its sheer size. Further research could study if accuracy on large and diverse datasets also benefits from multi-domain training.

\section{Conclusion} \label{sec:conclusion}
From the results it can be concluded that accuracy on individual domains can be improved by training on multiple domains, if batch size is taken into account to equalize hardware performance between all models, as hardware resources are not limitless. This means that semantic segmentation accuracy on a domain improves if the model is trained on additional data from a different domain, even when the domains have nothing in common. The multi-domain model has a 70\% larger output label-space than the largest baseline model, uses the same memory during training and has 2.5 times faster inference than the baseline models combined.\\

\bibliographystyle{IEEEtran}
\bibliography{references}
\end{document}